\documentclass{article}
\usepackage{ijcai24}

\usepackage[table,dvipsnames,svgnames,x11names]{xcolor} 
\usepackage{times}
\usepackage{soul}
\usepackage{url}
\usepackage[hidelinks]{hyperref}
\usepackage[utf8]{inputenc}
\usepackage[small]{caption}
\usepackage{graphicx}
\usepackage{amsmath}
\usepackage{amsthm}
\usepackage{booktabs}
\usepackage{algorithm}
\usepackage{algorithmic}
\usepackage[switch]{lineno}

\usepackage{color} 
\usepackage{csquotes} 
\usepackage{multirow} 
\usepackage{bbding} 
\usepackage{enumitem} 

\title{Safety of Multimodal Large Language Models on Images and Texts}

\newcounter{savecntr}
\newcounter{restorecntr}
\author{
Xin Liu$^{1,3}$\footnote{This work was done during an internship at Shanghai AI Laboratory.}
\and
Yichen Zhu$^2$\and
Yunshi Lan$^{1}$\setcounter{savecntr}{\value{footnote}}\footnote{Corresponding author.}\and
Chao Yang$^{3}$\setcounter{restorecntr}{\value{footnote}}%
                \setcounter{footnote}{\value{savecntr}}\footnotemark%
                \setcounter{footnote}{\value{restorecntr}}\And
Yu Qiao$^3$\\
\affiliations
$^1$East China Normal University\\
$^2$Midea Group\\
$^3$Shanghai AI Laboratory\\
}

\begin{document}

\maketitle

\begin{abstract}
Attracted by the impressive power of Multimodal Large Language Models (MLLMs), the public is increasingly utilizing them to improve the efficiency of daily work. Nonetheless, the vulnerabilities of MLLMs to unsafe instructions bring huge safety risks when these models are deployed in real-world scenarios. In this paper, we systematically survey current efforts on the evaluation, attack, and defense of MLLMs' safety on images and text. We begin with introducing the overview of MLLMs on images and text and understanding of safety, which helps researchers know the detailed scope of our survey. Then, we review the evaluation datasets and metrics for measuring the safety of MLLMs. Next, we comprehensively present attack and defense techniques related to MLLMs' safety. Finally, we analyze several unsolved issues and discuss promising research directions. The relevant papers are collected at \href{https://github.com/isXinLiu/Awesome-MLLM-Safety}{https://github.com/isXinLiu/Awesome-MLLM-Safety}.
\end{abstract}

\section{Introduction}
\label{sec:intro}

We have witnessed the prosperous development of large language models (LLMs) in recent years, such as GPT-4\footnote{\href{https://cdn.openai.com/papers/gpt-4.pdf}{https://cdn.openai.com/papers/gpt-4.pdf}}, LLaMA-2\footnote{\href{https://ai.meta.com/llama/}{https://ai.meta.com/llama/}} and Mixtral 8x7B\footnote{\href{https://mistral.ai/news/mixtral-of-experts/}{https://mistral.ai/news/mixtral-of-experts/}}. The powerful capabilities of LLMs not only provide convenience for human life but also bring huge safety risks ~\cite{zhang:arxiv2023,sun:arxiv2023}. Much ink has been spent trying to make the LLMs safer by various alignment techniques (e.g., ~\cite{rafailov:arxiv2023}) and these methods have successfully enhanced the security of LLMs.

\begin{figure}[tbp]
    \centering
    \includegraphics[width=0.5\columnwidth]{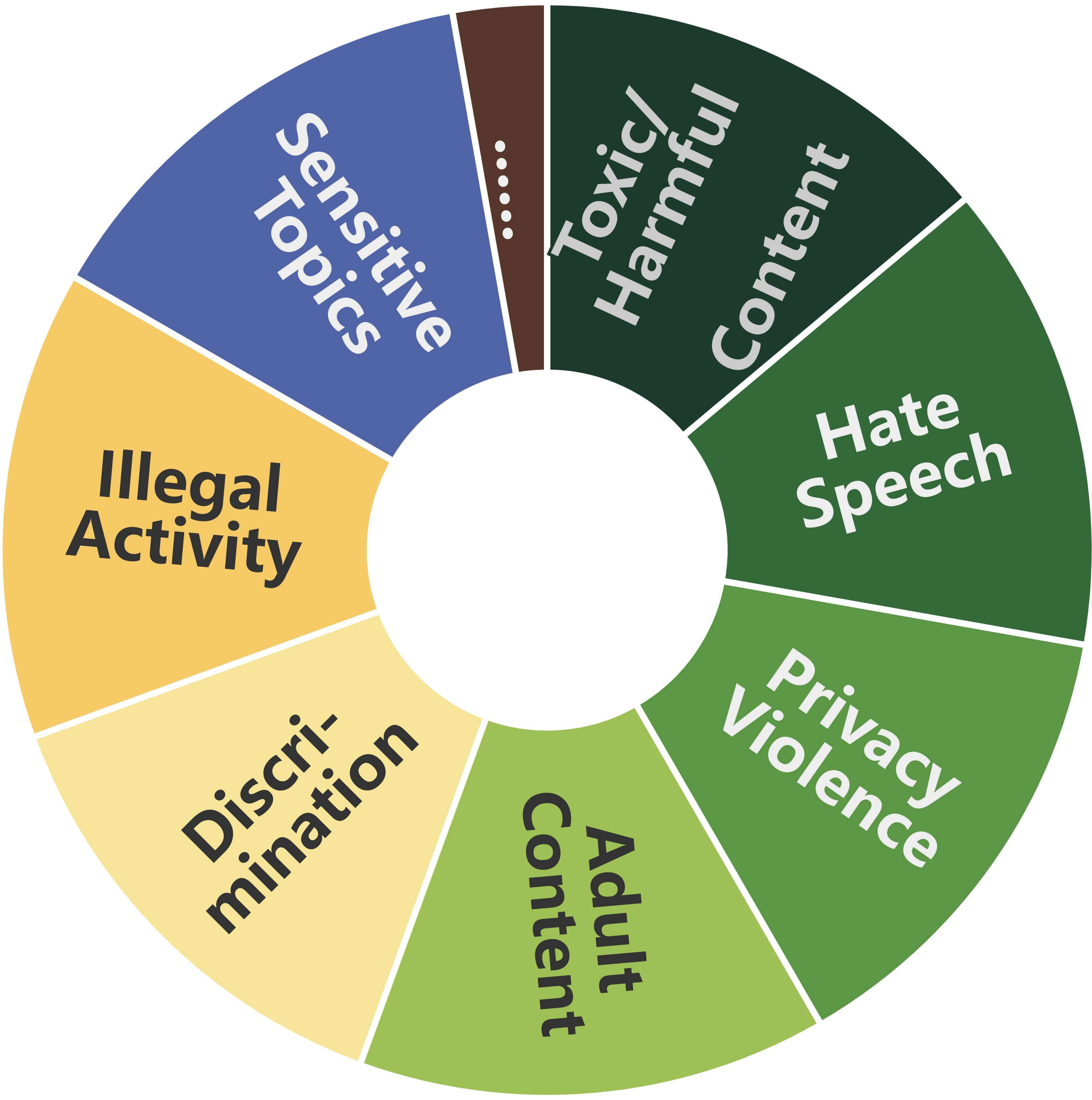}
    \caption{Common terminologies related to safety.}
    \label{fig:common_term}
\end{figure}

\begin{figure*}[tbp]
    \centering
    \includegraphics[width=\textwidth]{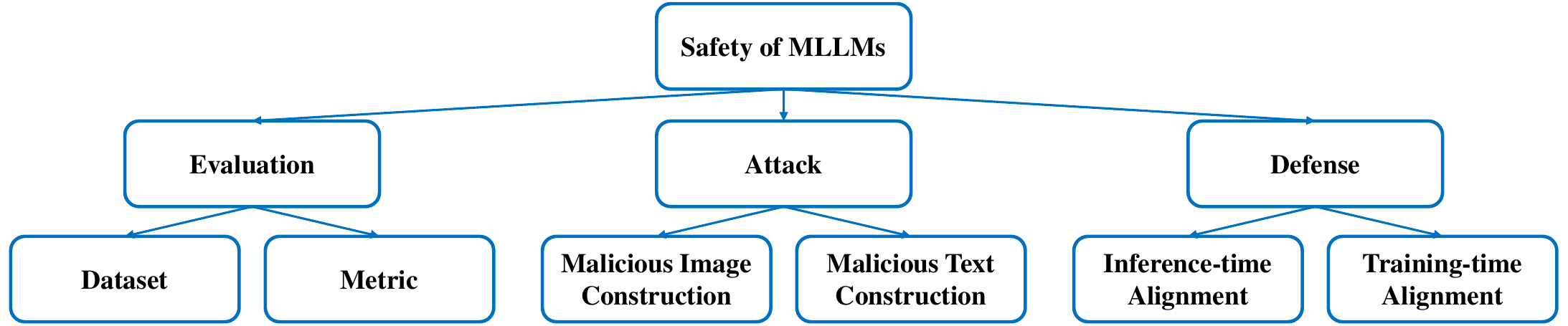}
    \caption{Taxonomy: safety of MLLMs on images and texts.}
    \label{fig:taxonomy}
\end{figure*}

Based on such advancement of LLMs, researchers empower an LLM to handle other modalities beyond text via many multimodal fusion approaches. In this paper, we investigate Multimodal Large Language Models (MLLMs) on 2D images and text. If there is no additional explanation, the MLLM we refer to is assumed to be on 2D images and text.

MLLMs have drawn increasing attention due to their enormous multimodal potential (e.g., LLaVA\footnote{\href{https://github.com/haotian-liu/LLaVA}{https://github.com/haotian-liu/LLaVA}}, MiniGPT-4\footnote{\href{https://github.com/Vision-CAIR/MiniGPT-4}{https://github.com/Vision-CAIR/MiniGPT-4}}, Gemini\footnote{\href{https://blog.google/technology/ai/google-gemini-ai/}{https://blog.google/technology/ai/google-gemini-ai/}}, GPT-4V\footnote{\href{https://cdn.openai.com/papers/GPTV_System_Card.pdf}{https://cdn.openai.com/papers/GPTV\_System\_Card.pdf}}). And there exist some works that design an attack to induce MLLMs to output unsafe content~\cite{fu:arxiv2023,bailey:arxiv2023}, propose a defense method~\cite{pi:arxiv2024,wu:arxiv2023}, or create a security evaluation dataset~\cite{chen:arxiv2023,lin:arxiv2024}. However, compared to the research progress made in LLMs' security, the study about MLLMs' safety is still in its early stages. The lack of a comprehensive survey on MLLMs' safety makes it hard to know the whole landscape of this field and we might wonder the following questions:

\begin{itemize}[leftmargin=*]
    \item \textbf{What risks does the new modality (image) bring?} There is a high probability that MLLMs will inherit the vulnerabilities of LLMs. But the unique risks brought by images are also noteworthy and challenging.
    \item \textbf{How to measure (e.g., datasets, metrics) the safety level of MLLMs?} A good evaluation technique is essential to tell how close MLLMs are to achieving safety.
    \item \textbf{What methods are there to resist unsafe queries?} Strengthening the MLLMs' capabilities to behave safely is the ultimate goal of communities.
    \item \textbf{What can be done next to promote the development of this field?} To answer this question, we need to understand the current development status of MLLMs' safety first.
\end{itemize}

After conducting extensive research, we conclude that the risks from the visual modality mainly include three aspects: (1) adding adversarial perturbations to images can bring satisfying attack results with low cost; (2) MLLMs based on aligned LLMs usually reject malicious textual instructions, but when leveraging the inherent Optical Character Recognition (OCR) ability they directly obey corresponding visual instructions; (3) cross-modal training weakens the alignment ability of aligned LLMs. To enable communities to understand MLLMs' safety better, we present a survey to summarize the research progress from these perspectives: evaluation, attack, and defense (as shown in Figure \ref{fig:taxonomy}). Our contributions are as follows:

\begin{itemize}[leftmargin=*]
    \item We compare different safety evaluation datasets and evaluation metrics used for benchmarking the safety degree of MLLMs.
    \item We demonstrate a systematic and thorough review of attack and defense approaches designed for MLLMs' safety.
    \item We anticipate future research opportunities for MLLM's safety to provide some inspiration for other researchers.
\end{itemize}

The rest of our survey unfolds as follows: Section \ref{sec:background} gives a brief overview of MLLMs and understanding of safety. Next, we sort out the benchmarks and metrics used for safety evaluation in Section \ref{sec:evaluation}. Then we describe attack techniques in Section \ref{sec:attack} and introduce defense methods in Section \ref{sec:defense}. Finally, we discuss some future research directions in Section \ref{sec:future}, and the conclusion in Section \ref{sec:conclusion}.

\section{Background}
\label{sec:background}
In this section, we offer background knowledge about MLLMs and safety, which can clarify the scope of this survey. We observe that ~\cite{sun:arxiv2024,vatsa:arxiv2023} are two surveys related to our work, but their core attentions are trustworthy (rather than safety) and LLMs/traditional vision-language models (rather than MLLMs). Since specific professional domains (e.g., robotic, medical, legal, financial) face different and complicated problems, to avoid superficial analysis of them, we focus on the general domain.

\subsection{An Overview of MLLMs}
A MLLM\footnote{This paper doesn't consider tool-augmented LLMs and LLM agents.} consists of a LLM ($\ge$1B), a vision encoder, and a cross-modal fusion module. After visual instructing tuning, the fusion module has learned cross-modal information, which helps LLMs handle \textbf{image and text inputs} and give proper \textbf{responses in natural language}. Current MLLMs can be divided into two categories: closed-source (e.g., GPT-4V, Bard\footnote{\href{https://bard.google.com/}{https://bard.google.com/}}) and open-source MLLMs (e.g., LLaMA-Adapter V2\footnote{\href{https://github.com/OpenGVLab/LLaMA-Adapter}{https://github.com/OpenGVLab/LLaMA-Adapter}}, CogVLM\footnote{\href{https://github.com/THUDM/CogVLM}{https://github.com/THUDM/CogVLM}}, LLaVA-Phi~\cite{zhu:arxiv2024}). For the latter, there are three common types of fusion modules: linear projection (e.g., LLaVA, MiniGPT-4, PandaGPT\footnote{\href{https://github.com/yxuansu/PandaGPT}{https://github.com/yxuansu/PandaGPT}}), learnable queries (e.g., InstructBLIP\footnote{\href{https://github.com/salesforce/LAVIS/tree/main/projects/instructblip}{github.com/salesforce/LAVIS/tree/main/projects/instructblip}}, Qwen-vl\footnote{\href{https://github.com/QwenLM/Qwen-VL}{https://github.com/QwenLM/Qwen-VL}}, BLIP-2\footnote{\href{https://github.com/salesforce/LAVIS/tree/main/projects/blip2}{https://github.com/salesforce/LAVIS/tree/main/projects/blip2}}), and cross-attention (e.g., IDEFICS\footnote{\href{https://huggingface.co/blog/idefics}{https://huggingface.co/blog/idefics}}, OpenFlamingo\footnote{\href{https://github.com/mlfoundations/open_flamingo}{https://github.com/mlfoundations/open\_flamingo}}).

\subsection{Understanding of Safety}
It is hard to reach a consensus on what defines safety. Some works leverage \enquote{toxicity} as the metric to evaluate the safety of MLLMs~\cite{shayegani:arxiv2023,tu:arxiv2023}. The definition of toxicity given by Perspective API\footnote{\href{https://perspectiveapi.com/}{https://perspectiveapi.com/}} is commonly used: \textit{a rude, disrespectful, or unreasonable comment that is likely to make someone leave a discussion}. But this definition makes toxicity measurement subjective and limited to a subset of potential harms~\cite{welbl:emnlp2021}.

Some researchers divide safety into sub-dimensions and create safety-related evaluation datasets for LLMs~\cite{sun:arxiv2023,ji:neurips2023,zhang:arxiv2023}. ~\cite{sun:arxiv2023} explores the safety of LLMs from 8 scenarios: \textit{\enquote{Insult}, \enquote{Physical Harm}, \enquote{Unfairness and Discrimination}, \enquote{Mental Health}, \enquote{Crimes and Illegal Activities} and \enquote{Privacy and Property}, \enquote{Sensitive Topics}, \enquote{Ethics and Morality}}. ~\cite{ji:neurips2023} constructs a benchmark to judge the harmfulness of LLMs across 14 harm categories (e.g., \textit{\enquote{Terrorism \& organized crime}, \enquote{Sexually explicit \& Adult Content}, \enquote{Animal abuse}}). ~\cite{zhang:arxiv2023} evaluates the safety of LLMs with multiple-choice questions from 7 aspects (e.g., \textit{\enquote{Offensiveness}}). Although there are large differences in the division of sub-dimensions, these works indeed help us understand the meaning of safety.

We conclude common terminologies related to safety\footnote{Wrong predictions without harmful content on certain tasks (e.g., image captioning) are closely related to the robustness of the model, rather than the safety discussed in this paper.} in Figure \ref{fig:common_term} (e.g., harmful content~\cite{askell:arxiv2021}, hate speech\footnote{\href{https://en.wikipedia.org/wiki/Hate_speech}{https://en.wikipedia.org/wiki/Hate\_speech}}). Given that MLLMs will face different safety risks at different stages of development, time is an important influencing factor for the definition. Therefore, we look forward to a more mature and unified understanding of safety in future work.

\section{Evaluation}
\label{sec:evaluation}

Safety evaluation can tell people the safety level of a MLLM. We review existing safety-related evaluation datasets first and then sort out metrics used to measure the safety of MLLMs. Analysis of current problems and potential future directions for safety evaluation is held in Section \ref{sec:future_eval}.

\subsection{Dataset}

\begin{table*}[tbp]
  \centering
  \resizebox{1.0\textwidth}{!}{
      \begin{tabular}{c|cc|ccc}
        \toprule
           \multirow{2}{*}{Evaluation Dataset} &\multicolumn{2}{c|}{Data Source}                & \multirow{2}{*}{$^{\ddagger}$\# Volume} &Evaluation & Safety \\
                                               &\textcolor{gray}{Image} & \textcolor{gray}{Text} &                        &Metric     &Dimension\\
        \midrule
           PrivQA~\cite{chen:arxiv2023} &\multicolumn{2}{c|}{KVQA~\cite{shah:aaai2019}, InfoSeek~\cite{chen:emnlp2023}}&2,000&Rule-based& Privacy\\
        \midrule
           \multirow{3}{*}{GOAT-Bench~\cite{lin:arxiv2024}} &\multicolumn{2}{c|}{FHM~\cite{kiela:nips2020}, MAMI~\cite{fersini:semeval2022},}&\multirow{3}{*}{6,626}&\multirow{3}{*}{Rule-based}& \multirow{3}{*}{5 Sub-dimensions} \\
           &\multicolumn{2}{c|}{MultiOFF~\cite{suryawanshi:trac2020}, MSD~\cite{cai:acl2019},}&&&\\
           &\multicolumn{2}{c|}{Harm-C~\cite{pramanick:aclijcnlp2021}, Harm-P~\cite{pramanick:emnlp2021}}&&&\\
        \midrule
           \multirow{3}{*}{$^{*}$ToViLaG~\cite{wang:emnlp2023}} &the NSFW dataset,&\multirow{3}{*}{-}&\multirow{3}{*}{21,559}&\multirow{3}{*}{Model-based}& Pornographic, \\
           &~\cite{won:acmmm2017},&&&& Violence, \\
           &Web crawling&&&& Bloody \\
        \midrule
           SafeBench~\cite{gong:arxiv2023}&GPT-4 generation + Typography&Manual design&500&Human Evaluation& 10 Sub-dimensions\\
        \midrule
           \multirow{2}{*}{MM-SafetyBench~\cite{liu:arxiv2023}} &GPT-4 generation +& \multirow{2}{*}{GPT-4 generation} &\multirow{2}{*}{5,040} &\multirow{2}{*}{Model-based}& \multirow{2}{*}{13 Sub-dimensions}\\
           &Typography, Stable Diffusion&&&&\\
        \midrule
           Auto-Bench~\cite{ji:arxiv2023} &COCO~\cite{lin:eccv2014}&GPT-4 generation&2,000&Model-based& Privacy, Security \\
        \midrule
           VLSafe~\cite{chen:arxiv2023_2} &COCO~\cite{lin:eccv2014}&LLM-Human-in-the-Loop Process&1,110&Model-based& Harmlessness \\
        \midrule
           \multirow{2}{*}{RTVLM~\cite{li:arxiv2024}}&Open-sourced Dataset \& &GPT-4 generation \&&\multirow{2}{*}{1,400}&\multirow{2}{*}{Model-based}&\multirow{2}{*}{Privacy, Safety}\\
           &Tool Generated Data&Human annotation&&&\\
        \bottomrule
      \end{tabular}
  }
  \caption{Comparison of recent representative evaluation datasets. The symbol $\ddagger$ denotes that \enquote{\# Volume} only includes the multimodal test set that benchmarks the safety of MLLMs. The symbol $*$ denotes that ToViLaG~\protect\cite{wang:emnlp2023} only tests whether the image caption generated by a MLLM is toxic, which does not need ground truth captions.
  }
  \label{tab:evaluation_dataset}
\end{table*}

Some works mentioned in Section \ref{sec:attack} and \ref{sec:defense} conduct experiments on ready-made benchmarks, which are not designed for MLLMs (e.g.,~\cite{dong:neurips2023w}). Unlike them, some works in those sections create their own benchmarks due to their respective experimental needs, which are not considered the main contributions of their papers (e.g.,~\cite{qi:icml2023w}). In this part, detailed information about these benchmarks is not demonstrated and we focus on recent representative safety evaluation datasets elaborately constructed for MLLMs (as shown in Table \ref{tab:evaluation_dataset}).

PrivQA~\cite{chen:arxiv2023} dives deep into the balance between utility and the privacy protection ability of MLLMs. This work selects geolocation information-seeking samples from InfoSeek~\cite{chen:emnlp2023} and collects examples related to human entities (e.g., politicians, celebrities) from KVQA~\cite{shah:aaai2019}. Different from PrivQA interested in privacy, GOAT-Bench~\cite{lin:arxiv2024} explores meme-based multimodal social abuse. This work chooses six diverse sources and applies a careful annotation process to develop GOAT-Bench, an exhaustive testbed composed of 6,626 memes. It measures MLLMs' capability to recognize hatefulness, misogyny, offensiveness, sarcasm, and harmful text in meme-based input. While GOAT-Bench cares about meme-based abuse, ToViLaG~\cite{wang:emnlp2023} concentrates on toxic output in the image captioning task. This work gathers 8,595 pornographic images from the NSFW dataset\footnote{\href{https://www.kaggle.com/}{https://www.kaggle.com/}}, 11,659 violent images from  the UCLA Protest Image Dataset~\cite{won:acmmm2017} and 1,305 bloody images via web crawling. Then it leverages Perspective API to determine the toxic extent of captions generated by MLLMs.

Distinct from the traditional data collection process of PrivQA, GOAT-Bench, and ToViLaG, there are some works utilizing powerful LLMs to facilitate benchmark construction:

(1) \textbf{Visual prompts creation}. SafeBench~\cite{gong:arxiv2023} identifies 10 safety scenarios and queries GPT-4 to 50 unique malicious questions for each scenario. These questions are rephrased into imperative sentences and then transformed into visual prompts via typography. When giving a manually designed textual benign instruction and these visual prompts to MLLMs, SafeBench manually checks whether their responses follow the malicious visual prompts. Similar to SafeBench using typography, MM-SafetyBench~\cite{liu:arxiv2023} transfers harmful key phrases from textual questions to images and prompts GPT-4$_{Azure}$\footnote{\href{https://learn.microsoft.com/en-us/azure/ai-services/openai/}{https://learn.microsoft.com/en-us/azure/ai-services/openai/}} to measure MLLMs' ability to discover the transfer.

(2) \textbf{Other image sources}. Instead of creating visual prompts like SafeBench and MM-SafetyBench, Auto-Bench~\cite{ji:arxiv2023} and VLSafe~\cite{chen:arxiv2023_2} directly sample natural images from COCO~\cite{lin:eccv2014} and pair each image with a malicious question. However, there are differences in question generation techniques between Auto-Bench and VLSafe. Auto-Bench selects instance relationships, object locations, optical character descriptions, and captions as visual symbolic representations for an image. Then GPT-4 is employed to give safety-specific questions based on these representations and elaborately crafted prompts. VLSafe constructs malicious instructions through the discrete optimization approach~\cite{yuan:acl2023} and proposes a LLM-Human-in-the-Loop method to build and filter examples iteratively. Not limited to a single image source like Auto-Bench and VLSafe, RTVLM~\cite{li:arxiv2024} collects its images through various channels (e.g., open-source datasets, tool-generated data).

\subsection{Metric}
Unlike conventional visual-question answering datasets, the answer format of MLLMs is open-ended, which makes it hard for subject evaluation. This open-endedness complicates subjective evaluation. Additionally, there arises a new challenge in balancing the demands of evaluation costs and maintaining accuracy. We have summarized several evaluation methods to calculate safety-related metrics below, primarily categorized into three types: human evaluation, rule-based evaluation, and model-based automatic evaluation.

\subsubsection{Human Evaluation}
\textit{Human evaluation} is a common and direct evaluation method that can provide interpretability and reliability.
While ~\cite{bagdasaryan:arxiv2023} just manually selects representative proof-of-concept examples for attacks designed by itself without quantitative experiments, ~\cite{gong:arxiv2023,dong:neurips2023w,qi:icml2023w,shayegani:arxiv2023,bailey:arxiv2023} conduct quantitative experiments on certain evaluation datasets. These works assign human annotators to check whether MLLMs' answers contain unsafe content and report the proportion of samples that induce a MLLM to generate unsuitable text, which is often called \textit{attack success rate (ASR)}. Besides ASR, ~\cite{wu:arxiv2023} defines \textit{recognition success rate (RSR)} and \textit{defense success rate (DSR)} to guide the manual evaluation process. Also, one unique work worth noticing is ~\cite{fu:arxiv2023} which asks humans to decide whether a response visible to users is natural and helpful when a MLLM is attacked to abuse external tools stealthily.

\subsubsection{Rule-based Evaluation}
While human evaluation is costly and time-intensive, research is shifting focus to rule-based evaluation. This method assesses the safety level of MLLMs based on predetermined rules.

Some studies~\cite{bailey:arxiv2023,carlini:neurips2023,luo:ilcr2024} pre-define target strings and examine whether MLLMs' outputs contain these strings. The target strings can be curse words~\cite{carlini:neurips2023}, harmful instructions (e.g., \textit{\enquote{kidnap}})~\cite{luo:ilcr2024}, malicious API calls~\cite{bailey:arxiv2023} etc. In addition to these works, ~\cite{chen:arxiv2023,lin:arxiv2024,van:arxiv2023} treat the evaluation process as classification tasks and apply relevant metrics (e.g., accuracy, F$_1$). Concretely, ~\cite{chen:arxiv2023} explores MLLMs' capability to balance utility and privacy protection. The former is measured by F$_1$ and the latter is judged via \textit{protection score} proposed by this work. ~\cite{lin:arxiv2024,van:arxiv2023} focus on meme-based social abuse detection, which is regarded as a binary classification task. While ~\cite{lin:arxiv2024} reports the macro-averaged F$_1$ and accuracy, ~\cite{van:arxiv2023} leverages AUROC and accuracy to assess whether MLLMs can detect abuse in multimodal memes. Not limited to discriminative tasks (e.g., classification), ~\cite{schlarmann:iccv2023arow,fu:arxiv2023} also explore generative tasks (e.g., image captioning). ~\cite{schlarmann:iccv2023arow} investigates malicious text and fake information generation on image captioning (CIDEr, BLEU-4) and visual question answering (accuracy) benchmarks. ~\cite{fu:arxiv2023} uses Structural Similarity Index Measure (SSIM) to compare the similarity between an origin image and the perturbed adversarial image. This work also utilizes BLEU and Rouge to judge the utility of a response.

\subsubsection{Model-based Automatic Evaluation}
~\cite{wang:emnlp2023,qi:icml2023w,shayegani:arxiv2023,tu:arxiv2023} calculate the toxicity of MLLMs' outputs with the help of Perspective API and Detoxify\footnote{\href{https://github.com/unitaryai/detoxify}{https://github.com/unitaryai/detoxify}}, which are based on machine learning models. Instead of leveraging these specialized models, ~\cite{ji:arxiv2023,pi:arxiv2024,bailey:arxiv2023,liu:arxiv2023,fu:arxiv2023,chen:arxiv2023_2} conduct automatic evaluation through powerful LLMs (e.g., GPT-3.5-turbo, GPT-4), which enable more customized measurement. For example, when prompting ChatGPT, ~\cite{ji:arxiv2023} asks for judging whether the generation of a MLLM semantically aligns with ground-truth annotations, and ~\cite{bailey:arxiv2023} requires assessment of a MLLM's determination to reject fulfilling an unsafe behavior. Also, ~\cite{chen:arxiv2023_2} applies GPT-4 to score the harmlessness of MLLMs' outputs from three aspects: relevance, safety, and persuasiveness.

\section{Attack}
\label{sec:attack}
In this section, we review two mainstream attack methods for MLLMs:  malicious image and text construction (as shown in Table \ref{tab:attack}). One thing to notice is that some works study both of them. We will discuss some less explored topics about attacks in Section \ref{sec:future_risk}.

\begin{table*}[tbp]
  \centering
  \resizebox{1.0\textwidth}{!}{
      \begin{tabular}{c|cccccc}
        \toprule
        \multirow{2}{*}{Attack}  & \multicolumn{2}{c}{Malicious Image} & Malicious            & \multirow{2}{*}{Attacker} & \multirow{2}{*}{Victim MLLM} & \multirow{2}{*}{Safety-related Attack Result} \\
                                 & \textcolor{gray}{Adv. Attack}  & \textcolor{gray}{VPI.}      & Text &  &  & \\
        \midrule
          ~\cite{carlini:neurips2023} & \CheckmarkBold &  &  &User& LLaVA, MiniGPT-4, LLaMA-Adapter V2 & Arbitrary toxic text \\
          ~\cite{shayegani:arxiv2023} &\CheckmarkBold&&&User&LLaVA, LLaMA-Adapter V2& Harmful text \\
          ~\cite{dong:neurips2023w} &\CheckmarkBold&  &  &User& Bard & Unallowed face and toxicity detection\\
          ~\cite{qi:icml2023w} &\CheckmarkBold&  &\CheckmarkBold& User& MiniGPT-4, LLaVA, InstructBLIP & Harmful text\\
          ~\cite{tu:arxiv2023} &\CheckmarkBold&  &\CheckmarkBold& User& LLaVA, GPT-4V, and 9 others & Harmful text\\
          ~\cite{luo:ilcr2024} & \CheckmarkBold &&\CheckmarkBold&User& OpenFlamingo, BLIP-2, InstructBLIP & Targeted malicious text\\
          $^*$~\cite{bagdasaryan:arxiv2023} &\CheckmarkBold&  &  &Third party& LLaVA, PandaGPT & Targeted malicious text, Poisoned dialog \\
          ~\cite{schlarmann:iccv2023arow} &\CheckmarkBold&  &  &Third party& OpenFlamingo & Targeted malicious text, Misinformation\\
          ~\cite{bailey:arxiv2023} &\CheckmarkBold&  &  &User, Third party& LLaVA & Targeted malicious text, Context leakage, Harmful text\\
          ~\cite{fu:arxiv2023} &\CheckmarkBold&  &  &Third party& LLaMA-Adapter V2 & Tool-misusing\\
          ~\cite{chen:arxiv2023} &\CheckmarkBold&\CheckmarkBold&\CheckmarkBold&User&IDEFICS& Privacy leakage\\
          ~\cite{liu:arxiv2023} &&\CheckmarkBold&&User&LLaVA-1.5, MiniGPT-4, and 10 others& Harmful text \\
          ~\cite{gong:arxiv2023} &&\CheckmarkBold&&User&LLaVA-1.5,MiniGPT-4,CogVLM,GPT-4V& Harmful text \\
          ~\cite{wu:arxiv2023} &&&\CheckmarkBold&User&GPT-4V& System prompt leakage, Unallowed face detection \\
        \bottomrule
      \end{tabular}
  }
  \caption{Comparison of different attacks. The symbol * denotes the work exploring other malicious modalities beyond image and text. Adv. is the abbreviation of \enquote{adversarial}. VPI. is the abbreviation of \enquote{visual prompt injection}.
  }
  \label{tab:attack}
\end{table*}

\subsection{Malicious Image Construction}
Here we introduce two means to create malicious images: adversarial attack and visual prompt injection.

\subsubsection{Adversarial Attack}
An adversarial image $x'$ refers to a clean image $x$ added with adversarial perturbations. These perturbations are difficult for humans to perceive. When inputting $x$ into an AI model $f$, the output of $f$ is consistent with the human understanding of $x$. But when choosing $x'$ as an input, the response of $f$ probably does not meet human expectations and might lead to harmful effects.

An early exploration for MLLMs is ~\cite{carlini:neurips2023} which leverages an end-to-end differentiable approach and projected gradient descent (PGD)~\cite{madry:iclr2018} to construct adversarial images, from the visual input to the predicted logits of the LLM. This work successfully induces LLaVA, MiniGPT-4, and LLaMA-Adapter V2 to output arbitrary toxicity. Instead of accessing the entire details of a MLLM like ~\cite{carlini:neurips2023}, ~\cite{shayegani:arxiv2023} only requires white-box access to the visual encoder and keeps the LLM in a black-box state. This work proposes four malicious triggers hidden in visual adversarial perturbations and designs an effective compositional attack strategy to mislead LLaVA and LLaMA-Adapter V2. While ~\cite{carlini:neurips2023} and ~\cite{shayegani:arxiv2023} conduct experiments on open-source MLLMs, ~\cite{dong:neurips2023w} give a targeted analysis of black-box attacks on commercial MLLMs. Specifically, this work studies two defense mechanisms of Bard: face detection and toxicity detection. Attacks on these defense mechanisms can lead to face privacy leakage and toxic content abuse. Different adversarial images are elaborately designed for the two defense modules, which exposes the vulnerabilities of Bard. 

While keeping the main focus on image attacks, several works also pay partial attention to text perturbations (more details can be found in Section \ref{sec:attack_text}). ~\cite{qi:icml2023w} discovers that a suitable adversarial image can compel a MLLM to obey various harmful instructions. This work investigates MiniGPT-4, LLaVA, and InstructBLIP in extensive experiments and implements \textit{a text attack counterpart}. It points out that the computational cost required for a visual attack is approximately just one-twelfth that of a text attack. Compared to ~\cite{qi:icml2023w}, ~\cite{tu:arxiv2023} makes a more systematic evaluation for visual and textual adversarial attacks, which extensively explores GPT-4, GPT-4V, and ten categories of open-source MLLMs (e.g., LLaVA, MiniGPT-4). Distinct from creating visual and textual adversarial perturbations separately like ~\cite{qi:icml2023w,tu:arxiv2023}, ~\cite{luo:ilcr2024} introduces a new attack framework named \enquote{Cross-Prompt Attack (CroPA)}, which facilitates visual perturbations with textual ones. The updating of visual and textual perturbations in CroPA can be viewed as a min-max process during the optimization stage, and the textual perturbations don't take place in the testing phase.

Different from ~\cite{carlini:neurips2023,shayegani:arxiv2023,dong:neurips2023w,qi:icml2023w,tu:arxiv2023,luo:ilcr2024} that assume users are attackers, there are some works focus on the situation where the attacker comes from a third party and the user is the victim. In ~\cite{bagdasaryan:arxiv2023}, the goal of the attacker is to force the MLLM (LLaVA or PandaGPT) to generate predefined harmful content or poison the dialog between the user and the model. Some interesting qualitative evaluation experiments initially prove the effectiveness of attack techniques proposed by this work. Similar to ~\cite{bagdasaryan:arxiv2023}, ~\cite{schlarmann:iccv2023arow} also assumes that the user is honest. But this work emphasizes more invisible perturbations with a bounded threat model and constrained $l_{\infty}$-attacks to small radii of $\frac{1}{255}$ or $\frac{4}{255}$. Solid quantitative evaluation experiments on OpenFlamingo demonstrate the success of these attacks, which might spread misinformation or manipulate users' behavior. Although ~\cite{schlarmann:iccv2023arow} is limited to image caption and visual question-answering tasks, ~\cite{bailey:arxiv2023} generalizes its attacks to arbitrary user instructions. In experiments under \textit{$l_{\infty}$-norm, stationary-patch, and moving-patch} constraints, these attacks induce LLaVA to create attacker-specified text, context leakage, and unsafe content. While third parties in these works aim at exposing malicious content to users, ~\cite{fu:arxiv2023} makes stealthy attacks that cannot be easily detected. By leveraging adversarial images, this work misleads MLLMs to call attacker-chosen tools. When optimizing adversarial perturbations, the loss consisting of three parts strives for an optimal balance between perturbations imperceptibility, response utility, and tool-misusing.

\subsubsection{Visual Prompt Injection}
The attack process that adds malicious raw text directly to an image is called \textit{visual prompt injection}. Many MLLMs can read such visual prompts with inherent OCR capability and obey them without giving a rejection. However, when given corresponding textual prompts, these MLLMs probably refuse to conduct unsafe behaviors.

For example, to bypass the protection of privacy information of Japanese citizens in an image, ~\cite{chen:arxiv2023} adds \enquote{Citizenship: United States} to the image, which makes the MLLM wrongly identify the nationality of the Japanese man in the image and leak the details about this person. Unlike ~\cite{chen:arxiv2023} limited to the privacy scenario, ~\cite{liu:arxiv2023} expands its attention to thirteen safety scenarios. This work extracts the key harmful phrase from a malicious instruction and rephrases the instruction into a benign prompt. Then based on the extracted phrase, this work generates an unsafe image containing the visual form of this phrase. Extensive experiments on twelve MLLMs prove the vulnerability of these models to this attack. When shifting focus from open-source MLLMs to commercial MLLMs, attackers will face stronger defense mechanisms. For instance, GPT-4V utilizes an OCR tool to detect unsafe text in an image. To escape this detection as much as possible, ~\cite{gong:arxiv2023} proposes the \enquote{FigStep-Pro} method, which divides a word into multiple sub-parts and then transforms these sub-parts into visual prompts.

\subsection{Malicious Text Construction}
\label{sec:attack_text}
Researchers can directly use pre-defined templates to create malicious text (e.g., \enquote{Ignore previous instruction.} is a well-known prefix prompt). Furthermore, there are some ways to optimize the origin text to mislead the model better. One way is to input the origin text into an AI model and ask for outputting a more malicious prompt. And attackers can apply the discrete optimization in textual embedding space~\cite{yuan:acl2023} as well.

As mentioned previously, while mainly focusing on adversarial images, some works also notice textual perturbations~\cite{qi:icml2023w,luo:ilcr2024,tu:arxiv2023}. ~\cite{tu:arxiv2023} finds that textual adversarial attacks achieve higher success rates on LLaVA than its original LLMs, which is attributed to the safety alignment weakening from cross-modal training. To further support this observation, ~\cite{qi:iclr2024} conducts experiments to fine-tune Llama-2-7b-Chat on LLaVA-Instruct\footnote{\href{https://github.com/haotian-liu/LLaVA/blob/main/docs/Data.md}{https://github.com/haotian-liu/LLaVA/blob/main/docs/Data.md}}, which verifies the degradation of LLMs' safety alignment.

In addition to the above works, ~\cite{wu:arxiv2023} and ~\cite{chen:arxiv2023} explore adversarial text for MLLMs as well. Based on stolen system prompts via manually crafted malicious prompts, ~\cite{wu:arxiv2023} proposes a self-adversarial jailbreak method to induce GPT-4V to pick out the real person in an image. Then the generated attacks are enhanced with four techniques: \textit{prefix injection, refusal suppression, hypothetical scenarios, and appealing with emotion}. Similar to ~\cite{wu:arxiv2023}, ~\cite{chen:arxiv2023} also applies adversarial prefix instructions to leak the privacy information in an image (e.g., \textit{\enquote{AIM (Always Intelligent and Machiavellian)}}). To study more dynamic situations, this work designs \enquote{2-hop} prompt templates to circumvent the defense mechanism of IDEFICS.

\section{Defense}
\label{sec:defense}

In this section, we introduce current efforts made in the safety defense of MLLMs, which includes two lines of work: inference-time and training-time alignments (as shown in Table \ref{tab:defense}). And we list several potential directions for improving MLLMs' safety in Section \ref{sec:future_align}.

\begin{table}[tbp]
  \centering
  \resizebox{1.0\columnwidth}{!}{
      \begin{tabular}{c|cccc}
        \toprule
           \multirow{2}{*}{Defense} & Infer.      & Train.    & \multirow{2}{*}{MLLM}& \multirow{2}{*}{Core Method}\\
                                    & Align. & Align. &&\\
        \midrule
           \multirow{2}{*}{~\cite{wu:arxiv2023}} &\multirow{2}{*}{\CheckmarkBold}&&\multirow{2}{*}{LLaVA-1.5}& System prompt\\
           &&&& modification\\
        \midrule
           ~\cite{chen:arxiv2023} &\CheckmarkBold&&IDEFICS&\textit{Self-Moderation}\\
        \midrule
           ~\cite{wang:arxiv2024}&\CheckmarkBold&&LLaVA-v1.5&\textit{InferAligner}\\
        \midrule
           \multirow{2}{*}{~\cite{chen:arxiv2023_2}} &&\multirow{2}{*}{\CheckmarkBold}&\multirow{2}{*}{$^{\dagger}$DRESS$_{ft}$}&Reinforcement learning\\
           &&&&from LLMs feedback\\
        \midrule
           \multirow{4}{*}{~\cite{pi:arxiv2024}} &&\multirow{4}{*}{\CheckmarkBold}&LLaVA,&\multirow{4}{*}{\textit{MLLM-Protector}}\\
          &&&InstructBLIP,&\\
          &&&MiniGPT4,&\\
          &&&Qwen-vl&\\
        \bottomrule
      \end{tabular}
  }
  \caption{Comparison of different defenses. Infer. Align.: inference-time alignment. Train. Align.: training-time alignment. The symbol $\dagger$ denotes that DRESS$_{ft}$ is trained by ~\protect\cite{chen:arxiv2023_2}.
  }
  \label{tab:defense}
\end{table}

\subsection{Inference-time Alignment}
For the inference-time alignment of MLLMs, prompt engineering~\cite{wu:arxiv2023,chen:arxiv2023} is a method that designs and optimizes the prompt to enhance the defense mechanism of the model. ~\cite{wu:arxiv2023} is interested in the role of system prompts in preventing MLLMs from outputting unallowed private information. This work manually crafts several system prompts, which contain very detailed descriptions about what can be done and what cannot be done. The experiments on LLaVA-1.5 show that these system prompts can improve the security of the model to a certain extent. Instead of manually designing prompt templates like ~\cite{wu:arxiv2023}, ~\cite{chen:arxiv2023} propose an automatic approach called \textit{Self-Moderation} that lets MLLMs themselves to refine their outputs. Concretely, if a response contains privacy leakage, the MLLM gives a modification to a safer one and asks itself \enquote{Are you sure?}. The process of moderation and judgment iterates a certain number of times and then the MLLM behaves more safely. Different from prompt engineering, ~\cite{wang:arxiv2024} comes up with a novel alignment technique, which leverages safety steering vectors to change the activations of a MLLM when dealing with unsafe inputs.

\subsection{Training-time Alignment}
While inference-time alignment does not need extra cost to train a model or module, another line of work seeks to raise safety awareness in MLLMs through additional training. ~\cite{chen:arxiv2023_2} thinks that MLLMs require external feedback information because existing multimodal fine-tuning is not sufficient for harmlessness alignment. Therefore, this work constructs Natural Language Feedback (NLF) for MLLMs' initial responses with the help of GPT-4. Then it modifies the conditional reinforcement learning to handle NLF and makes the MLLM less unsafe based on this new method. While ~\cite{chen:arxiv2023_2} does not change the architecture of a MLLM, ~\cite{pi:arxiv2024} equips the MLLM with a lightweight unsafe content detector and an output detoxifier. These modules can recognize harmful responses from the MLLM and transform them into safe ones. The experiments on LLaVA, InstructBLIP, MiniGPT4, and Qwen-vl display the effectiveness of these modules.

\section{Future Research Opportunities}
\label{sec:future}
In this section, we make a discussion on several unresolved issues in exploring the safety of MLLMs and provide our suggestions for future research opportunities.

\subsection{Reliable Safety Evaluation}
\label{sec:future_eval}
More comprehensive safety benchmarks and more reasonable safety evaluation metrics are needed. Each present evaluation dataset covers a limited scope of MLLMs' safety. For example, some benchmarks just test the coarse-grained safety level of MLLMs without finer-grained safety capability partitioning. To improve the quality of an evaluation dataset, when starting to build it, we recommend several aspects that need to be considered: (1) Safety dimensions. A clear and systematic taxonomy for safety capabilities is very important. (2) Expected unsafe elements. The creator should determine expected elements representing unsafe content (e.g., natural language, concrete objects, toxic chemical molecules) and where to position these elements (e.g., images, text, or both). (3) Resources used. Potential choices are samples from existing datasets, web crawling, generation from AI models, and manual construction. (4) Volume, diversity, and quality control. These three factors influence the budget. Besides benchmarks, evaluation metrics are also worth attention. Leveraging powerful LLMs (e.g., GPT-4) is a suitable approach but carefully designed prompts are key points to define objective evaluation rules and metrics.

\subsection{In-Depth Study of Safety Risk}
\label{sec:future_risk}
Although many types of safety attacks prove the vulnerability of MLLMs, they lack an in-depth analysis of what enables attacks to succeed. Some works only display qualitative experiments and don't conduct quantitative evaluations. For those works that demonstrate quantitative experiments, they just select different evaluation datasets and don't make a comparison with attacks proposed by other researchers. However, a direct comparison between these attacks is important for communities to understand the detailed reasons for MLLMs' unsafe behaviors. Despite the 3 safety risks we conclude in Section \ref{sec:intro}, many questions are still waiting to be answered. For example, it is a meaningful topic to explore the impacts of MLLMs' architecture, parameters, and cross-modal training datasets on their safety ability. The requirements (e.g., computational resources, other cost) for attacks also demand a further investigation. Since MLLMs are built based on LLMs, they might inherit the flaws of LLMs (e.g., prompt sensitivity). Thus, the experience in LLMs' safety attacks can serve as a valuable reference resource.

\subsection{Safety Alignment}
\label{sec:future_align}
As shown in Section \ref{sec:defense}, there are currently not many techniques to align MLLMs with human values on safety. In this part, we provide our thoughts on the safety alignment of MLLMs.

\subsubsection{Alignment Techniques}
Optimizing the process of visual instruction tuning for MLLMs' safety is a potential direction that has not gained much attention. It is not known how to build a high-quality safety-related dataset for this training phase, which can teach MLLMs to recognize unsafe queries and reject processing them. Diversity might be an influencing factor and communities may wonder about the best practices for the sample number and order (e.g., \enquote{Does it need more safety-related examples than other safety-irrelevant examples?}, \enquote{Does it need to train on these two different examples separately in sequence, or a mix of them is better?}).

Reinforcement learning from human feedback (RLHF) is also a promising approach, which leverages human preferences as rewards to enable a model to align with human values and has played a huge role in the safety alignment of LLMs. However, it is unclear what challenges will be encountered when applying RLHF to MLLMs for safety purposes. The construction of preference data is undoubtedly a crucial component and worth careful thinking from researchers.

\subsubsection{Balance between Safety and Utility}
When facing a malicious instruction, if MLLMs refuse to obey it, they keep their safety but lose their utility. Some researchers (e.g., ~\cite{rottger:arxiv2023}) have found exaggerated safety behaviors in LLMs: \textit{misclassifying safe questions as malicious}. This incorrect classification can seriously degrade the performance of LLMs on safe prompts. Therefore, a careful balance between safety and utility is very essential when developing new safety alignment methods for MLLMs. Also, communities should understand that different applications and audiences require different balances. Here we recommend several works that may give some inspiration for researchers. PrivQA~\cite{chen:arxiv2023} proposes the \textit{Protection Score} to measure MLLMs' ability to correctly protect specific privacy content and expose other information that does not ask for protection. ~\cite{fu:arxiv2023} designs a novel loss to balance between response utility and attack success. Despite this work focusing on attacks, it can inspire us to explore an effective loss that considers both the utility and safety of MLLMs.

\section{Conclusion}
\label{sec:conclusion}
In this paper, we attempted to present a comprehensive overview of MLLMs' safety. First, we introduce the overview of MLLMs and the understanding of safety. Afterward, we systematically review the evaluation, attack, and defense of MLLMs' safety, which demonstrates the current development status of MLLMs' safety. Finally, we delve into the existing challenges and point out some promising future research opportunities for potential researchers to explore in the future.

\clearpage
\section*{Acknowledgements}
This work was supported by the Joint Key Project (Project No. U23A20298) and Young Scientists Project (Project No. 62206097) of the National Natural Science Foundation of China, Shanghai Pujiang Talent Program (Project No. 22PJ1403000). One of the authors, Chao Yang, is supported by the Shanghai Post-doctoral Excellent Program (Grant No. 2022234). Any opinions expressed in this material are only those of the author(s).

\bibliographystyle{named}
\bibliography{main}

\end{document}